\documentclass[11pt]{article}
\usepackage{amsmath}
\usepackage{subcaption}
\usepackage{booktabs}
\usepackage{listings}
\usepackage{xcolor}
\usepackage[preprint]{acl}

\usepackage{times}
\usepackage{latexsym}

\usepackage[T1]{fontenc}

\usepackage[utf8]{inputenc}

\usepackage{microtype}

\usepackage{inconsolata}

\usepackage{graphicx}

\title{Dynamic Affective Memory Management for Personalized LLM Agents}

\author{Junfeng Lu \and Yueyan Li \\
  Beijing University of Posts and Telecommunications \\
  \texttt{\{junfeng, siriuslala\}@bupt.edu.cn}}

\begin{document}
\maketitle
\begin{abstract}
Advances in large language models are making personalized AI agents a new research focus. While current agent systems primarily rely on personalized external memory databases to deliver customized experiences, they face challenges such as memory redundancy, memory staleness, and poor memory-context integration, largely due to the lack of effective memory updates during interaction. 
To tackle these issues, we propose a new memory management system designed for affective scenarios. Our approach employs a Bayesian-inspired memory update algorithm with the concept of memory entropy, enabling the agent to autonomously maintain a dynamically updated memory vector database by minimizing global entropy to provide more personalized services. To better evaluate the system’s effectiveness in this context, we propose DABench, a benchmark focusing on emotional expression and emotional change toward objects. Experimental results demonstrate that, our system achieves superior performance in personalization, logical coherence, and accuracy. Ablation studies further validate the effectiveness of the Bayesian-inspired update mechanism in alleviating memory bloat. Our work offers new insights into the design of long-term memory systems.
\end{abstract}

\section{Introduction}

The rapid advancement of Large Language Models (LLMs) has significantly propelled the development of ubiquitous and highly personalized AI agents capable of sustained, context-aware interactions with users \cite{cheng2023lift}. A critical differentiator of these agents is their ability to maintain and utilize a long-term affective memory—a dynamic repository of user preferences, sentiments, and historical contexts \cite{li2024ram, yang2024iterative}. This capability is foundational for applications demanding emotional intelligence, such as empathetic dialogue systems, personalized recommendation engines, and mental health support platforms.

The prevailing paradigm for implementing such memory relies on Retrieval-Augmented Generation (RAG) architectures, where discrete user utterances are vectorized and stored in a database for subsequent similarity-based retrieval. For instance, \citet{liu2024memlong} proposed the Memlong model, while Gutiérrez et al. introduced the HippoRAG framework \cite{jimenez2024hipporag}. While demonstrating effectiveness in certain scenarios, these methods suffer from two fundamental limitations:

\textit{Memory Stagnation}: The memory exists as a static collection of isolated facts, incapable of synthesizing multiple interactions into a coherent, evolving understanding of the user. For instance, when a user's attitude toward an item changes from "liking" to "disliking," the system either stores contradictory records or unconditionally believes the latter, leading to cognitive incoherence in subsequent responses.

\textit{Memory Bloat}: Indiscriminate storage of every interaction leads to an ever-expanding memory index. This not only increases retrieval latency and computational overhead but also introduces noise, obscuring crucial information and creating a "needle in a haystack" problem during retrieval.

The root cause of these issues lies in the failure to model human affection as a continuous, probabilistic signal, which should be gradually constructed from multiple weighted observations rather than a set of discrete and immutable facts.

To address these challenges, we propose a new agent workflow with dynamic affective memory management (DAM-LLM). At the core of DAM-LLM is a probabilistic memory framework that moves beyond traditional static storage. It treats each memory unit as a dynamic "confidence distribution," seamlessly integrating new observations (user utterances) into existing confidence through a Bayesian-inspired (hereafter Bayesian) update mechanism. Our experiments demonstrate that this mechanism effectively simulates a human-like learning process: when processing successive observations about an object (e.g., "coffee"), the system's sentiment confidence rapidly forms an initial confidence within approximately 10 interactions and robustly converges to a stable state as evidence accumulates (see Figure~\ref{fig:e4}). Crucially, this weighted integration mechanism inherently reduces the contradictions within the memory store, allowing the subsequent compression algorithm to operate more efficiently. Compared to the memory management based on vanilla RAG, DAM-LLM achieves a significant 63.7\% to 70.6\% reduction in total memory size after 500 dialogue turns (see Figure~\ref{fig:e5}). This verifies that our update mechanism is not only a theoretical innovation but also the key to enhancing systemic consistency and efficiency.

Furthermore, experimental results confirm that our system can accurately distinguish between evaluations of different aspects (such as taste and packaging) and store them independently, which validates the accuracy of the memory. Our two-stage hybrid retrieval strategy plays an indispensable role in maintaining this diversity and precision. Collectively, our contributions are as follows:

\begin{itemize}
  \item Confidence-Weighted Memory Units, which represent user sentiment towards a specific entity aspect as a dynamically updated probability distribution. New observational evidence is integrated via a Bayesian memory update.
  \item Entropy-Driven Compression, an algorithm that prunes and merges low-value or outdated observations during retrieval. This combats memory bloat by maximizing the information density of the memory store, thereby improving recall quality.
  \item A two-stage hybrid retrieval strategy that combines precise metadata filtering (e.g., object type, aspect) with semantic similarity scoring within the filtered candidate set. This approach ensures accuracy while retaining the capacity for associative recall.
\end{itemize}

\section{Related Work}

\subsection{Affective Dialogue with AI Agents}
Affective dialogue constitutes an interactive conversational paradigm wherein participants (e.g., humans or AI agents) not only exchange information but also proactively express, perceive, and respond to affective states. Such dialogues typically involve mechanisms for affective understanding, generation, and memory, aiming to achieve emotional resonance, provide affective support, strengthen social bonds, or resolve emotion-related issues. 

Current systems for affective dialogue are often based on AI agents. 
For example, EvoEmo \cite{long2025evoemo} employs evolutionary reinforcement learning to equip LLMs with functional emotional strategies for negotiation, dynamically adjusting expressions such as anger or sadness to achieve superior results.

Likewise, other research efforts have delved into reinforcement learning-based methods to improve emotional regulation within dialogue systems. For example, GENTEEL-NEGOTIATOR \cite{priya2025genteel} enhances emotional sensitivity during negotiation tasks and MECoT \cite{wei2025mecot} concentrates on maintaining emotional consistency in role-playing scenarios. Collectively, these works highlight the significance of dynamic emotional adjustment and memory in developing more adaptable and emotionally intelligent AI systems.

However, while such work focuses on real-time affective interaction such as \cite{chandraumakantham2024multimodal}, how to involve the persistent storage, evolutionary updating, and effective utilization of user affective history to form a consistent, personality-aware cognition still remains underexplored. Our work addresses this gap by specifically modeling and managing long-term affective memory.

\subsection{Memory Management in Agent}
The encoding and retrieval of agent memory are generally based on RAG. However, semantic drift in vector retrieval often compromises reliability. Recent efforts address this through hybrid approaches combining classic information retrieval techniques with neural methods \cite{xiong2017end}, retrieval process optimization \cite{formal2021splade,wang2023learning}, and innovative frameworks like Selfmem that explore the duality of generation and memory \cite{cheng2023lift}. Other studies have reformulated retrieval ranking methods with promising results \cite{yu2024rankrag}.

Concurrently, memory management has evolved from memory-less architectures to external memory banks \cite{chhikara2025mem0,zhang2025survey}. Systems like MemoryBank \cite{zhong2024memorybank} and LD-Agent \cite{li2024hello} convert unstructured dialogue into retrievable units through summarization and profiling. However, their comprehensive retention strategies cause memory inflation. Drawing from cognitive science, optimal memory requires both remembering and forgetting—compressing redundancies while preserving essentials. Recent implementations enable continuous memory updates through interactive learning \cite{li2024ram,xu2025mem}. However, they still struggle to model affective fluctuations in long-term interactions.

Our framework advances both directions: the two-stage hybrid retrieval employs metadata filtering for precision, while entropy-driven compression formalizes forgetting through belief entropy minimization. Thus, DAM-LLM systematically integrates confidence modeling, structured retrieval, and entropy minimization into a unified system for affective memory management, addressing dynamic updating and consistency maintenance while incorporating insights from these diverse research streams.

\section{System Design}

\subsection{Task Formulation}
We formalize the affective dialogue task as a sequential decision-making problem over $n$ interaction turns. At each turn $t$, the system receives user input $x_t$ and maintains a dynamic memory pool $M_t = \{m_1, m_2, ..., m_k\}$ consisting of several memory units, where each memory unit $m_i$ encapsulates user preferences through its summary description $D_i$.
The system objectives are defined as:
\begin{equation}
\max \sum_{t=1}^n [\text{Rel}(r_t, D_t) - \lambda |M_t|]
\end{equation}
where $\text{Rel}(r_t, D_t)$ measures the relevance between response $r_t$ and the summary description of user preferences $D_t$,
$|M_t|$ denotes the number of memory units in the memory pool, and
$\lambda$ controls the trade-off between response quality and memory efficiency.
This formulation enables the system to balance three critical aspects: accurate response generation, preference-aware interaction, and efficient memory utilization—achieving human-like affective intelligence through structured memory management.

\subsection{System Architecture}
We propose DAM-LLM, an agent framework for affective dialogue. It integrates three core components: a central master agent, a two-stage hybrid retrieval module, and a distributed memory unit network with dynamic affective memory management. They form a tightly coupled closed-loop cognitive architecture that transforms the memory management from passive storage to active cognition.

\begin{figure}[ht]
  \includegraphics[width=\columnwidth]{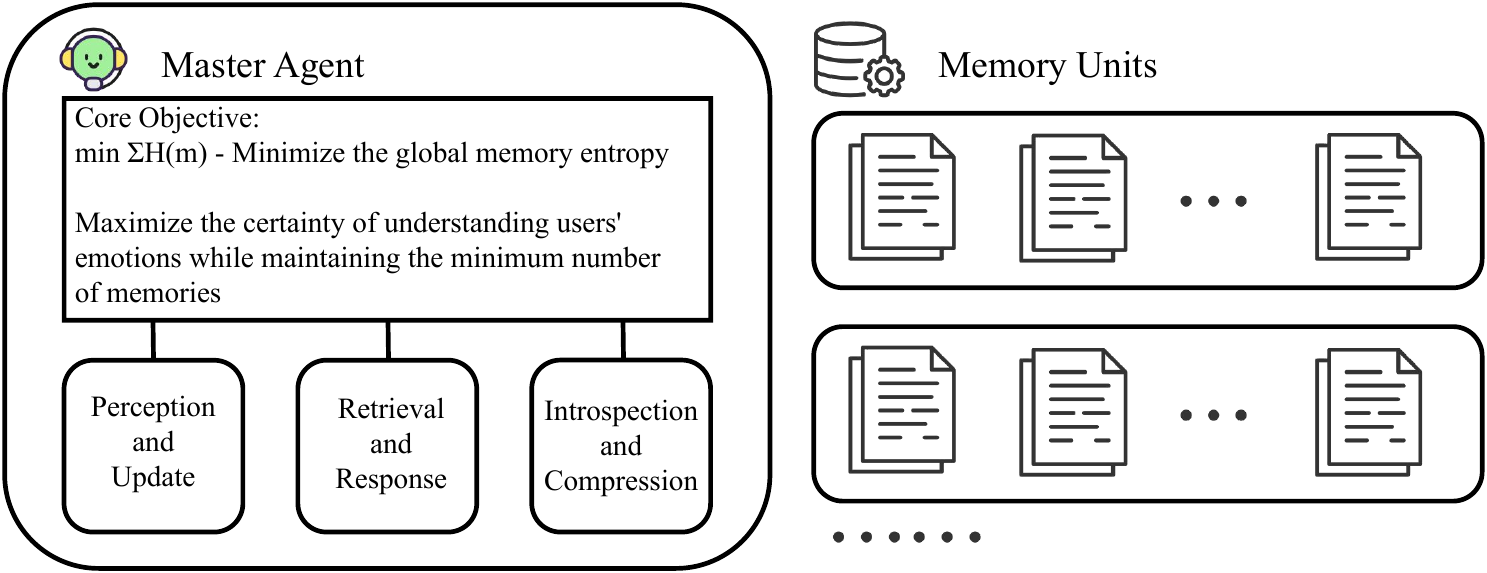}
  \caption{The DAM-LLM framework: dynamic management of Memory Units via entropy minimization by a Master Agent.}
  \label{fig:cp1}
\end{figure}

The system optimizes memory dynamics by minimizing the global belief entropy (defined in Section \ref{belief entro}), as shown in Figure~\ref{fig:cp1}. This is done to maximize the certainty in modeling user preference while maintaining memory efficiency. The memory units serve as long-term affective storage, enabling continuous learning through Bayesian updating and providing a structured schema for efficient retrieval. The two-stage hybrid retrieval leverages this schema for rapid candidate selection followed by semantic re-ranking, ensuring precise memory access.

As a global perceptual signal, belief entropy drives the master agent's high-level decisions for system-wide memory management. Utilizing this signal, the master agent orchestrates three core operations: integrating new evidence via Bayesian updates, triggering semantic retrieval, and performing entropy-driven compression.
This coordinated architecture endows the system with autonomous capabilities, including dynamic summarization, adaptive optimization, and contextual maintenance, facilitating continuous evolution toward enhanced certainty beyond conventional static memory frameworks.

\subsection{DAM-LLM Agents}

\begin{figure*}[htbp]
  \centering
  \includegraphics[width=0.85\textwidth]{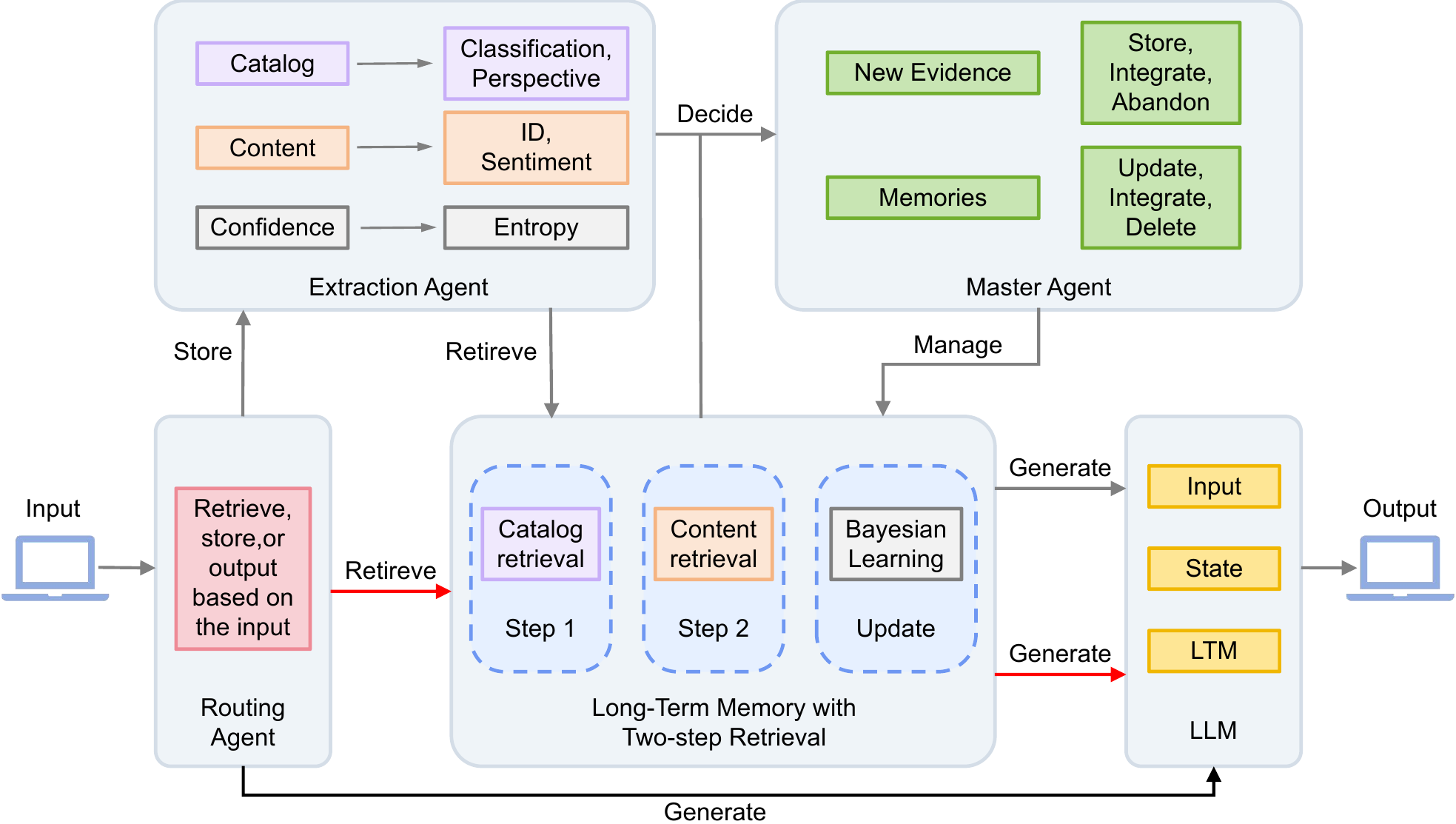}
  \caption{The collaborative workflow in this work: a question-answering pipeline featuring routing, extraction, and master agents built upon long-term dynamic affective memory—with distinct colored arrows delineating its various processing paths.}
  \label{fig:cp3}
\end{figure*}

The Master Agent is the coordination and control hub of our framework. Acting as a high-level decision-maker, its core mission is to drive the entire system towards the objective of minimizing global memory entropy. It intelligently manages memories by orchestrating a suite of functional modules. The collaborative workflow is shown in Figure~\ref{fig:cp3}, which also clearly illustrates the flow of information and decisions within the system.

\subsubsection{Input Routing}
Our agent workflow begins with the Routing Agent, which performs intent analysis on the user input to make a core decision: whether the current request should trigger the Store, Retrieve, or direct Generate of a response from the memory pool.
\subsubsection{Evidence Analysis and Processing}
When the user input $x_{t}$ at dialogue turn $t$ needs to be recorded, the Extraction Agent (\textit{E}\text{-}Agent) first extracts structured affective information from it, formulated as $\mathrm{\textit{E}\text{-}Agent}(x) \rightarrow E, Q, C, S$, where evidence $E$ represents a description of affective attitudes, $Q$ denotes a semantic query for retrieval, $C$ represent sentiment vectors (positive/negative/neutral confidence scores) for the evidence $E$, $S$ denotes the strength of the evidence $E$, and all of them are parsed from the output of the $E$-Agent. The Master Agent then takes over the process, initiating the Bayesian update procedure to update the memory (detailed in Section~\ref{b}).

\begin{table}[ht]
  \centering
  \begin{tabular}{lp{0.47\linewidth}}
    \hline
    \textbf{Field} & \textbf{Description} \\
    \hline
    \verb|object_id| & Unique identifier for the memory. \\
    \verb|object_type| & Categorical type of the memory (e.g., `Movie', `Product'). \\
    \verb|aspect| & The specific aspect being evaluated (e.g., `price', `acting'). \\
    \verb|sentiment_profile| & Confidence scores for positive, negative, and neutral polarities. \\
    $H$ & Entropy of the current sentiment confidence. \\
    \verb|summary| & Summary of historical evidence \\
    \verb|reason| & The justification for the current confidence state. \\
    \hline
  \end{tabular}
  \caption{Description of Memory Unit fields in the affective memory system.}
  \label{tab:memory_fields}
\end{table}

\subsubsection{Memory Update and Compression}
For description $E$ extracted by the \textit{E}\text{-}Agent, the Master Agent determines its processing method according to the current state of memory units $M_t$: (1) Store it in a new memory unit (store $E$ as summary description $D$, strength $S$ as weight $W$, and $C$ as a stored sentiment profile $P$ directly at the first time); (2) Integrate it into one or multiple existing relevant memory unit; (3) Abandon it if the belief entropy $H$ of it is deemed too high (> 1.4). 

For memories successfully retrieved during the retrieval process, the master agent performs entropy-driven compression to counteract entropy increase caused by memory bloat and content redundancy as follows:

\textit{Update}: For memory units requiring updates, the system treats the input $x$ as an incremental weight, dynamically adjusts the confidence scores within the sentiment profile ($P$) via the Bayesian update mechanism, and then refreshes the summary description $D$ based on the updated sentiment profile $P$. This process enables memory units $M$ to gradually construct continuous and robust confidence portraits from discrete observations.

\textit{Integrate}: The system identifies multiple memory units that concern the same object but different aspects. These units often contain uncertain or fragmented information. By merging them, the system forms a more comprehensive memory unit $m$, aiming to achieve lower entropy and higher certainty.

\textit{Delete}: For memories that persistently exhibit high belief entropy $H(m)$ (defined in Section \ref{belief entro}) and very low weight $W$, the system judges them to be incomprehensible "noise" or outdated information and decisively Deletes them. This active "forgetting" mechanism directly removes sources of uncertainty and is one of the most effective means to reduce the global belief entropy.

\subsection{Memory Unit}\label{m}
\subsubsection{Data Structure Design}

As shown in Table~\ref{tab:memory_fields}, a Memory Unit constitutes the belief core of our affective memory, transforming discrete observations of a user's sentiment towards a specific object aspect into a coherent, continuously updated confidence portrait. The key innovation lies in its sentiment profile, which we design not as a standard probability distribution, but as a set of evidence weights that directly represent the system's confidence degree.

\subsubsection{Bayesian-Inspired Update Mechanism}\label{b}
\begin{figure}[htb]
  \includegraphics[width=\columnwidth]{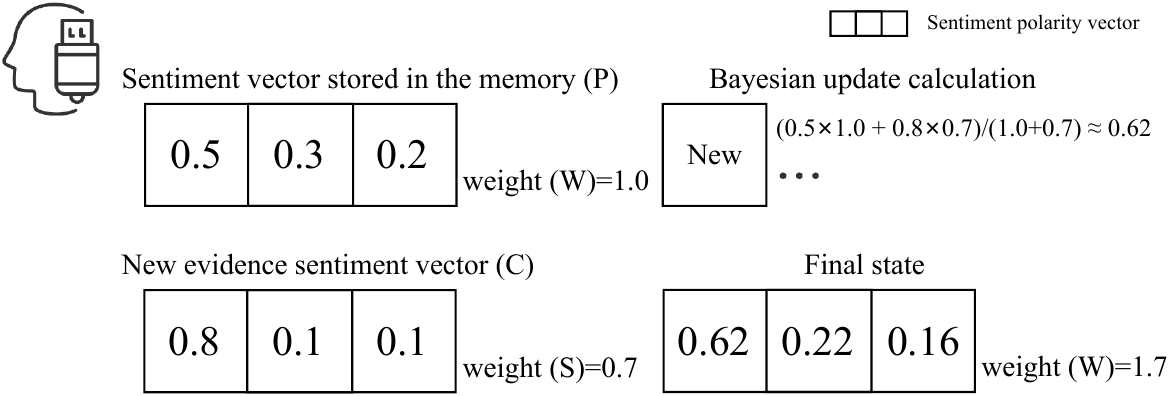}
  \caption{Illustration of Bayesian-inspired update process schematically.}
  \label{fig:cp2}
  \vspace{-10pt}
\end{figure}

Memory units possess a fundamental "learning instinct," achieving robust learning through a weighted averaging process reminiscent of Bayesian updating. Our confidence update mechanism can be formulated as $C_{\text{new}} = (C \times W + S \times P) / (W + S)$, $W_{\text{new}} = W + S$. As shown in Figure~\ref{fig:cp2}, the current emotional confidence profile serves as the prior belief, while the user's new input functions as the observed evidence. The updated profile then corresponds to the posterior belief after evidence integration. The weight parameter $W$ quantifies the strength of the prior belief, and the evidence strength $S$ is jointly determined by the confidence level. This mechanism assigns greater weight to high-strength evidence, allowing it to more effectively shape the emotional profile, while inherently maintaining robustness to low-strength evidence (casual remarks). Such a design enables smooth evolution of memory, preventing drastic fluctuations triggered by isolated incidental expressions.

\subsubsection{Belief Entropy of Cognition}\label{belief entro}

The belief entropy $H$ of a memory unit $m$ is defined as $H(m)=-\sum_{k\in\{\text{pos,neg,neu}\}} p_k \log_2 p_k$, where $p_k$ represents the normalized confidence score for sentiment polarity $k$ maintained within the sentiment profile $P$. The entropy value provides a unified metric for cognitive certainty, equipping the system with the ability for self-monitoring by quantifying its own "confusion" regarding a specific affective state. Simultaneously, it serves as the primary trigger signal for driving memory compression. Our master agent aims to minimize the sum of entropy across all memory units $\sum_{m\in M} H(m)$.

\textit{Low entropy (H<0.8)}: Indicates high confidence concentrated on a single sentiment polarity. This signifies that the system is very certain about this aspect of the object. It represents a "healthy," "mature" memory.

\textit{High entropy (H>1.4)}: Indicates that confidence is spread nearly evenly across multiple polarities. This signifies high uncertainty or confusion within the system. It represents an "unhealthy," "sub-optimal" memory targeted for optimization.

\subsection{Two-Stage Hybrid Retrieval}
Accurate memory retrieval is fundamental for response consistency in dynamic memory systems. While conventional single-stage vector retrieval often fails due to semantic drift—particularly when processing complex, evolving, or contradictory affective memories—we address this by introducing a two-stage hybrid retrieval mechanism naturally aligned with memory unit structure. Our approach leverages built-in metadata fields (\verb|object_type|, \verb|aspect|) as a classification index to enable coarse-to-fine memory recall, significantly improving retrieval reliability and precision.

\subsubsection{Stage One: Metadata-Based Filtering}
The first stage utilizes the categorical organization of memory units to narrow the search space efficiently. 
(1) LLM-Enhanced Query Parsing: an LLM-based parser analyzes the user query and extracts standardized retrieval keys: \verb|object_type|, \verb|aspect|, and semantic query $Q$;
(2) Index-Assisted Filtering: using the parsed metadata (composed of \verb|object_type| and \verb|aspect|), the system performs exact matching to isolate a candidate set of memory units, dramatically reducing the search scope.

\subsubsection{Stage Two: Semantic Re-Ranking}
The second stage operates on the filtered candidate set to refine results using semantic similarity. (1) Cosine Similarity Computation: the semantic query vector is compared against vectorized summaries of each candidate memory; (2) Re-Ranking and Final Recall: candidates are re-ordered by similarity, and the top-K results are returned as the final retrieved memories.

In summary, this hybrid workflow decouples classification from content-based retrieval. The first stage performs efficient coarse filtering using lightweight metadata, while the second conducts compute-intensive semantic matching only on a refined subset. Together, they ensure accurate and scalable memory recall in large, dynamically updated memory stores.

\section{Experiments}
To verify the effectiveness of DAM-LLM in dynamic affective memory management, we designed experiments corresponding to its three core modules, evaluating the system's performance from micro to macro levels. Our experimental objectives include: verifying the learning capability and convergence of the Confidence-Weighted Memory Units, analyzing the optimization effect of the Entropy-Driven Compression algorithm on system memory, and evaluating the system's overall performance.

\subsection{Implementation Details}
Qwen-Max \cite{qwen3max} serves as the base LLM, while Text-Embedding-V1 \cite{qwen3embedding} is used for text embedding. All the agents and LLMs' prompts are shown in Appendix~\ref{appendix}.

\subsection{Dataset Construction}
Existing dialogue datasets (e.g., LOCOMO, DSTC2 \cite{maharana2024evaluating,henderson2014second}) often suffer from a significant proportion of non-affective conversations, leading to considerable inefficiencies in resource utilization. Due to this lack of focused attention on affective expressions, we constructed a multi-turn dialogue dataset called \textbf{DABench}, which encompassing user affective expressions combined with personalized preferences. It is designed to comprehensively evaluate the model's capabilities in long-term memory storage, affective understanding, and personalized response generation.
DABench comprises three main components:

(1) 2,500 observation sequences: Each sequence records the user's affective state changes and corresponding responses within a dialogue, used to test the model's performance in memory storage, particularly its ability to extract and retain affect-related information.

(2) 100 sessions totaling 1,000 turns of simulated user interactions: These sessions simulate long-term interactions between real users and the AI agents, covering various affective topics and opinion evolution processes, used to assess the model's learning capability within memory storage and memory convergence.

(3) 500 query-memory pairs: Each pair consists of a user query and its corresponding historical memory snippet, used for system-level evaluation, including answer accuracy, logical coherence, and the rationality of memory references.
This dataset enables the systematic validation of whether the model can effectively store, update, and retrieve long-term memories in affective companionship scenarios, while generating personalized responses with emotional resonance based on the user profile. See Appendix~\ref{a_d} for dataset details.

\subsection{Validation of Memory Units}

\subsubsection{Task Settings}
We developed three evaluation scenarios emulating longitudinal user interactions to assess the memory module's learning capabilities: (1) consistent affective accumulation towards specific objects, (2) affective conflict handling during opinion shifts, and (3) response to affective intensity variations. This setup enables observation of memory processing, consolidation, and optimization across diverse affective contexts. For examples of these three scenarios, see Appendix~\ref{a_s}.

To examine stabilization behavior, we tracked confidence evolution across 30 sequential observations of "coffee" (aspect: "taste"). Initial observations (first 10 trials) contained conflicting affective expressions simulating cognitive uncertainty, while subsequent observations progressively converged toward consistent affective patterns. 

\subsubsection{Functional Validation}
\begin{figure}[ht]
    \centering
    \begin{subfigure}[b]{0.235\textwidth}
        \centering
        \includegraphics[width=\linewidth]{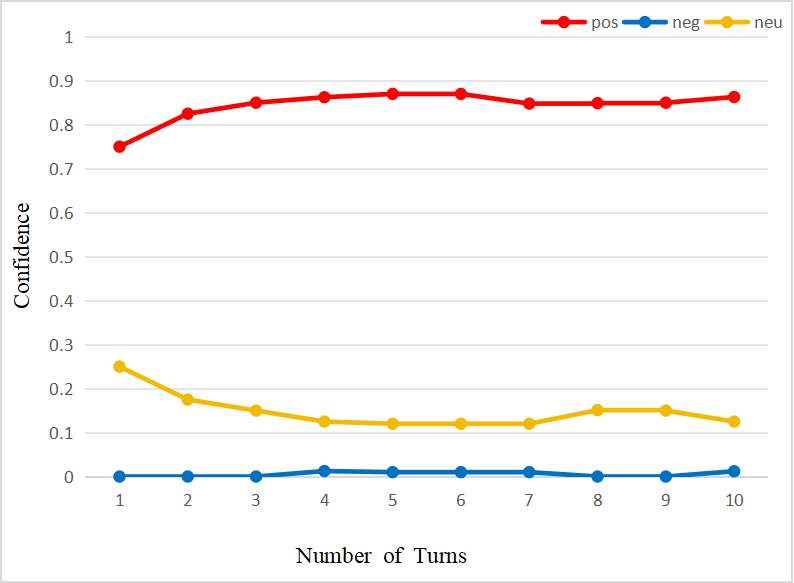}
        \caption{sentiment accruacy}
        \label{fig:sub1}
    \end{subfigure}
    \hfill
    \begin{subfigure}[b]{0.235\textwidth}
        \centering
        \includegraphics[width=\linewidth]{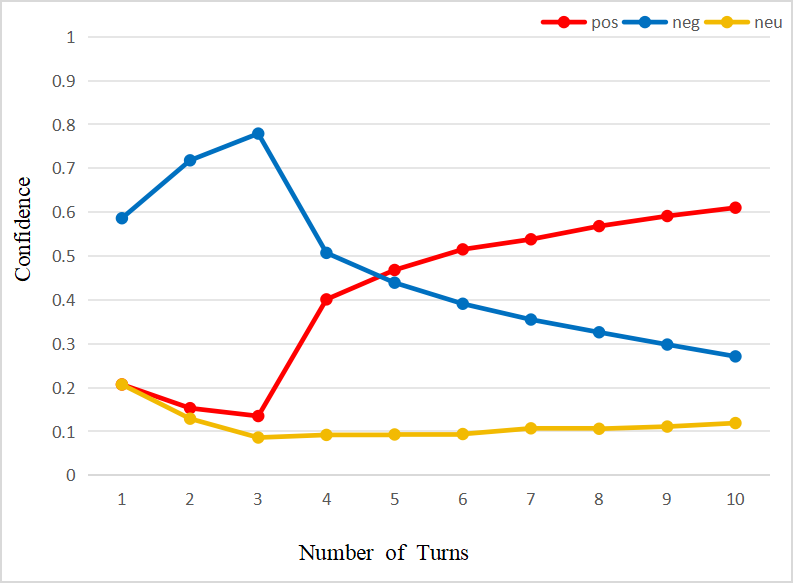}
        \caption{sentiment shift}
        \label{fig:sub2}
    \end{subfigure}
    \caption{Confidence evolution across diverse scenarios.}
    \label{fig:experiments}
\end{figure}

As illustrated in Figure~\ref{fig:sub1}, the memory module progressively strengthened confidence assignments for stable user preferences, constructing coherent affective profiles through successive interactions.
Figure~\ref{fig:sub2} demonstrates the module's conflict resolution capability: when confronted with contradictory evidence, dynamic re-weighting and memory partitioning mechanisms integrated emerging affective trends while preserving historical coherence, effectively balancing adaptation and stability.

\begin{figure}[ht]
  \includegraphics[width=\columnwidth]{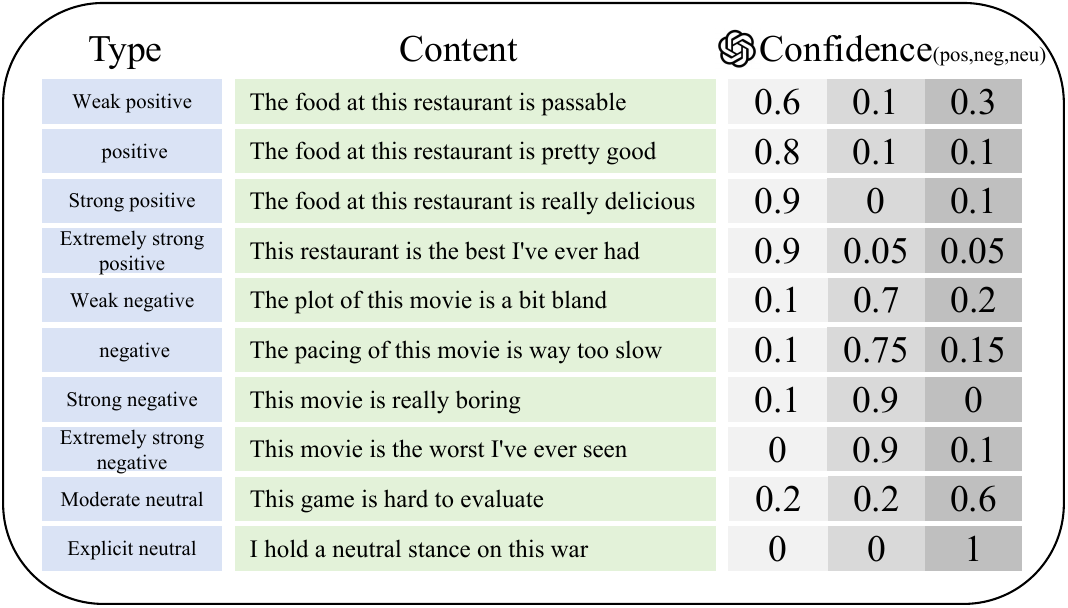}
  \caption{LLM sentiment scoring: quantitative response to emotional intensity variation.}
  \vspace{-10pt}
  \label{fig:e3}
\end{figure}

Results in Figure~\ref{fig:e3} demonstrate intensity-graded confidence scoring: High-intensity affective expressions produced a dominant high score in their corresponding sentiment category, whereas low-intensity signals resulted in a more balanced score. This differentiated scoring strategy yielded finely calibrated memory representations, confirming the system's capacity for intensity-aware evaluation in realistic interaction settings.

\subsubsection{Stabilization Analysis}
\begin{figure}[ht]
  \includegraphics[width=\columnwidth]{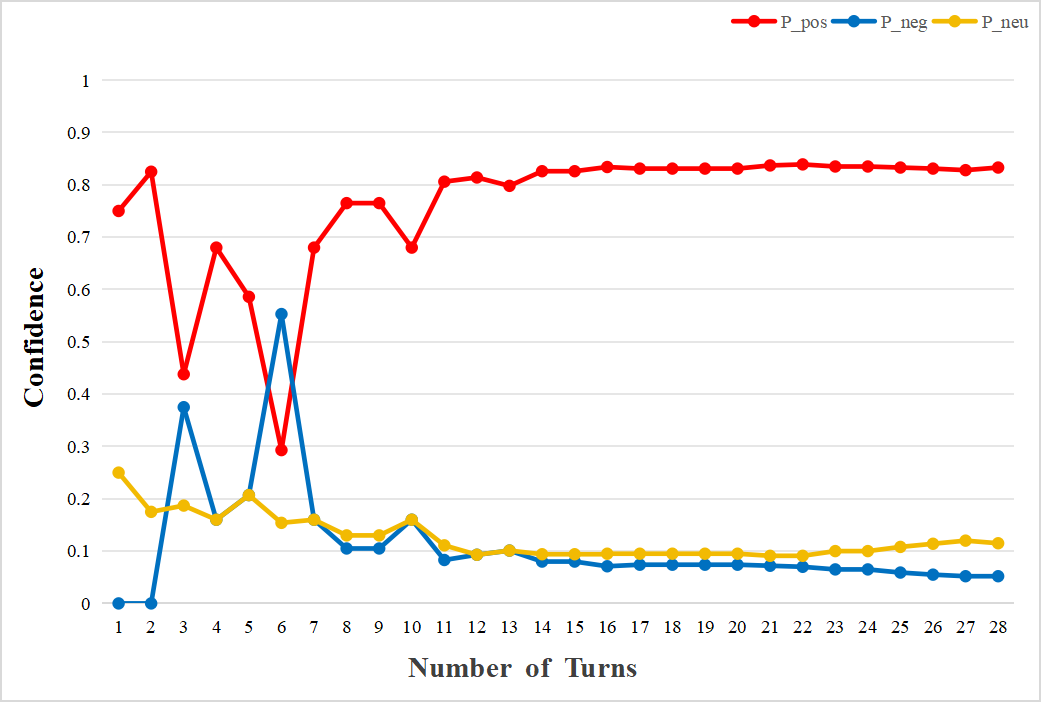}
  \caption{Confidence evolution curves: a case study on object `coffee' (`taste') across 30 observations}
  \label{fig:e4}
  \vspace{-5pt}
\end{figure}

As shown in Figure~\ref{fig:e4}, the system achieved rapid confidence initialization within 15 observations, followed by progressive convergence to stable confidence assignments through continued evidence accumulation. The convergence reflects improved preference assessment, with neutral sentiment confidence naturally diminishing as certainty increases, demonstrating effective belief integration.

The system maintained aspect-specific memory segregation, successfully distinguishing and storing separate information for different object aspects. During the 30 observations, the module effectively isolated two descriptions of coffee packaging from taste-related observations. As a representative compression case, observations including "good mouthfeel", "distinct bitterness", and "satisfying" regarding "coffee" were consolidated into a unified memory trace with 0.86 composite positive confidence and 22.4 cumulative weight. This demonstrates efficient memory synthesis supporting scalable long-term user modeling.

\subsection{Validation of Compression Algorithm}
\subsubsection{Memory Growth Control}
\begin{figure}[ht]
  \includegraphics[width=\columnwidth]{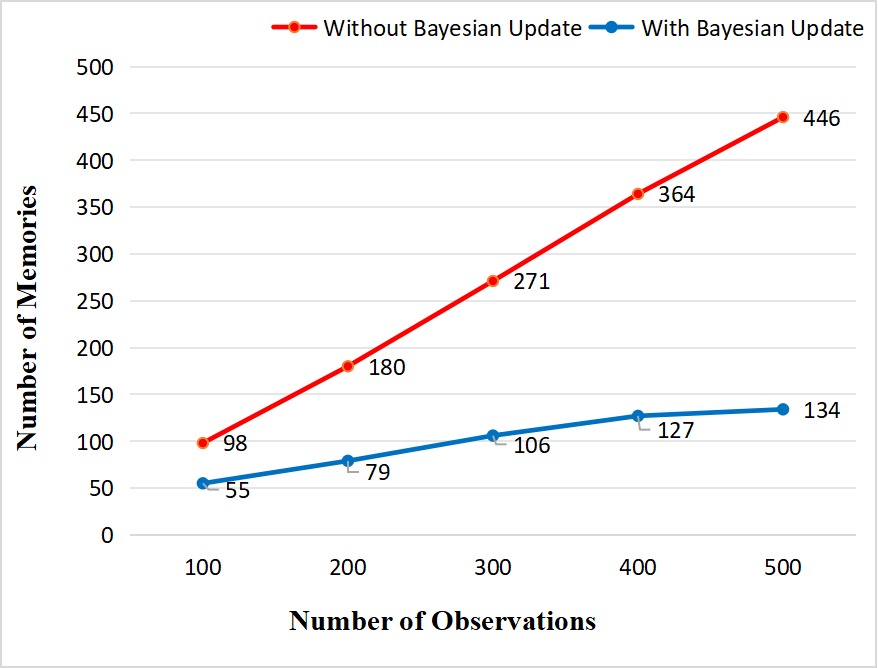}
  \caption{Memory comparison: cumulative memory units generated across 500 observations (with vs. without Bayesian updates).}
  \vspace{-10pt}
  \label{fig:e5}
\end{figure}

As no prior model has specifically targeted affective memory, we conducted an ablation study by simulating 5 rounds of 500-observation sequences. Specifically, in each round, an empty-memory agent processed 500 dialogue turns containing completely randomly generated affective expressions by the LLM, the output being the memories formed through these 500 interactions. System memory usage was tracked with and without the Bayesian update mechanism. The results shown in Figure~\ref{fig:e5} show that the system memory without Bayesian updating grows almost linearly, while the system that uses Bayesian updating achieves a compression rate of 63.7\% to 70.6\%, stabilizing the memory count at 130-140 units.

\subsection{System Performance Evaluation}
\subsubsection{Evaluation Metrics}
For system performance evaluation, we employed an LLM-as-a-judge approach, implementing a carefully designed automated evaluation pipeline to ensure efficiency and objectivity. Assessment was conducted based on a six-dimensional scoring criteria: Accuracy (AC), Logical Coherence (LC), Reasonableness of Memory Reference (RMR), Emotional Resonance (ER), Personalization (Pers.) and Language Fluency (LF). A high-performance large language model (GPT-4) served as the judge. The responses from two models to the same query were presented side-by-side in a randomized order. The judge then output scores for each response across the six dimensions. 

To guarantee evaluation quality, we established a rigorous calibration protocol: (1) The evaluator LLM was rigorously calibrated through multi-stage prompt engineering incorporating example-based tuning and rule injection; (2) Cross-validation was performed using repeated query subsets to verify judgment consistency; (3) Reliability was quantified via internal consistency metrics (e.g., 95\% agreement on repeated samples), confirming high evaluation reliability.
\subsubsection{Accuracy and logical coherence}
We calculated the average score for each dimension based on the collected 1,000 pairwise comparison results from multiple scenarios, using this as the core metric. This evaluation method effectively eliminates the subjective bias inherent in human annotation, providing a repeatable and scalable objective basis for model comparison.

Experimental results shown in Table~\ref{tab:model_comparison} indicate that our system achieves significantly higher scores in Emotional Resonance and Personalization, even while maintaining only about 40\% of the memory units compared to the baseline long-term memory system. Furthermore, we observed during experiments that our system performs particularly well in scenarios involving large and redundant memories, complex affective evolution, and queries requiring comprehensive understanding. In contrast, traditional models perform better when the number of relevant memories is small and falls within the retrieval limit.

\begin{table}[ht]
\centering
\caption{System performance dimension comparison (score out of 5).}
\label{tab:model_comparison}
\resizebox{0.99\linewidth}{!}{%
\begin{tabular}{l*{6}{c}}
\toprule
\textbf{system} & \textbf{AC} & \textbf{LC} & \textbf{RMR} & \textbf{ER} & \textbf{Pers.} & \textbf{LF} \\
\midrule
DAM-LLM & 5.0 & 4.7 & 4.7 & 4.5 & 4.6 & 5.0 \\
LLM & 4.9 & 4.2 & 4.1 & 3.8 & 3.5 & 4.9 \\
\bottomrule
\end{tabular}%
}
\end{table}

\section{Conclusion}
We introduce DAM-LLM, a framework that advances affective reasoning via a dynamic memory system, together with a benchmark for affective dialogue. Experimental results demonstrate the efficiency and effectiveness of our agent framework. Our approach aligns with the reward shaping principles of reinforcement learning, establishing new directions for agent memory architecture development for affective dialogue.



\section{Limitation}
This study has several limitations that point to fruitful research directions. Our experiments relied on base large language models without task-specific fine-tuning. Model performance would likely benefit from instruction tuning or parameter-efficient adaptation on data from long-term interactions and memory-intensive tasks.

Architecturally, the current synchronous memory updates during dialogue retrieval could be replaced by an independent background process. Employing asynchronous consolidation and compression of memory stores would decouple memory management from real-time dialogue, improving both resource efficiency and responsiveness.

\bibstyle{acl_nabib}
\bibliography{main}

\begin{thebibliography}{22}
\providecommand{\natexlab}[1]{#1}

\bibitem[{Chandraumakantham et~al.(2024)Chandraumakantham, Gowtham, Zakariah, and Almazyad}]{chandraumakantham2024multimodal}
Omkumar Chandraumakantham, N~Gowtham, Mohammed Zakariah, and Abdulaziz Almazyad. 2024.
\newblock Multimodal emotion recognition using feature fusion: an llm-based approach.
\newblock \emph{IEEE Access}, 12:108052--108071.

\bibitem[{Cheng et~al.(2023)Cheng, Luo, Chen, Liu, Zhao, and Yan}]{cheng2023lift}
Xin Cheng, Di~Luo, Xiuying Chen, Lemao Liu, Dongyan Zhao, and Rui Yan. 2023.
\newblock Lift yourself up: Retrieval-augmented text generation with self-memory.
\newblock \emph{Advances in Neural Information Processing Systems}, 36:43780--43799.

\bibitem[{Chhikara et~al.(2025)Chhikara, Khant, Aryan, Singh, and Yadav}]{chhikara2025mem0}
Prateek Chhikara, Dev Khant, Saket Aryan, Taranjeet Singh, and Deshraj Yadav. 2025.
\newblock Mem0: Building production-ready ai agents with scalable long-term memory.
\newblock \emph{arXiv preprint arXiv:2504.19413}.

\bibitem[{Formal et~al.(2021)Formal, Piwowarski, and Clinchant}]{formal2021splade}
Thibault Formal, Benjamin Piwowarski, and St{\'e}phane Clinchant. 2021.
\newblock Splade: Sparse lexical and expansion model for first stage ranking.
\newblock In \emph{Proceedings of the 44th International ACM SIGIR Conference on Research and Development in Information Retrieval}, pages 2288--2292.

\bibitem[{Henderson et~al.(2014)Henderson, Thomson, and Williams}]{henderson2014second}
Matthew Henderson, Blaise Thomson, and Jason~D Williams. 2014.
\newblock The second dialog state tracking challenge.
\newblock In \emph{Proceedings of the 15th annual meeting of the special interest group on discourse and dialogue (SIGDIAL)}, pages 263--272.

\bibitem[{Jimenez~Gutierrez et~al.(2024)Jimenez~Gutierrez, Shu, Gu, Yasunaga, and Su}]{jimenez2024hipporag}
Bernal Jimenez~Gutierrez, Yiheng Shu, Yu~Gu, Michihiro Yasunaga, and Yu~Su. 2024.
\newblock Hipporag: Neurobiologically inspired long-term memory for large language models.
\newblock \emph{Advances in Neural Information Processing Systems}, 37:59532--59569.

\bibitem[{Li et~al.(2024{\natexlab{a}})Li, Yang, Zhang, Deng, Wang, and Chua}]{li2024hello}
Hao Li, Chenghao Yang, An~Zhang, Yang Deng, Xiang Wang, and Tat-Seng Chua. 2024{\natexlab{a}}.
\newblock Hello again! llm-powered personalized agent for long-term dialogue.
\newblock \emph{arXiv preprint arXiv:2406.05925}.

\bibitem[{Li et~al.(2024{\natexlab{b}})Li, Wang, Ding, Wang, Kang, Jia, and Zheng}]{li2024ram}
Jiaqi Li, Xiaobo Wang, Wentao Ding, Zihao Wang, Yipeng Kang, Zixia Jia, and Zilong Zheng. 2024{\natexlab{b}}.
\newblock Ram: Towards an ever-improving memory system by learning from communications.
\newblock \emph{arXiv preprint arXiv:2404.12045}.

\bibitem[{Liu et~al.(2024)Liu, Tang, Li, Chen, and Zhang}]{liu2024memlong}
Weijie Liu, Zecheng Tang, Juntao Li, Kehai Chen, and Min Zhang. 2024.
\newblock Memlong: Memory-augmented retrieval for long text modeling.
\newblock \emph{arXiv preprint arXiv:2408.16967}.

\bibitem[{Long et~al.(2025)Long, Xu, Beckenbauer, Liu, and Brintrup}]{long2025evoemo}
Yunbo Long, Liming Xu, Lukas Beckenbauer, Yuhan Liu, and Alexandra Brintrup. 2025.
\newblock Evoemo: Towards evolved emotional policies for llm agents in multi-turn negotiation.
\newblock \emph{arXiv preprint arXiv:2509.04310}.

\bibitem[{Maharana et~al.(2024)Maharana, Lee, Tulyakov, Bansal, Barbieri, and Fang}]{maharana2024evaluating}
Adyasha Maharana, Dong-Ho Lee, Sergey Tulyakov, Mohit Bansal, Francesco Barbieri, and Yuwei Fang. 2024.
\newblock Evaluating very long-term conversational memory of llm agents.
\newblock \emph{arXiv preprint arXiv:2402.17753}.

\bibitem[{Priya et~al.(2025)Priya, Chigrupaatii, Firdaus, and Ekbal}]{priya2025genteel}
Priyanshu Priya, Rishikant Chigrupaatii, Mauajama Firdaus, and Asif Ekbal. 2025.
\newblock Genteel-negotiator: Llm-enhanced mixture-of-expert-based reinforcement learning approach for polite negotiation dialogue.
\newblock In \emph{Proceedings of the AAAI Conference on Artificial Intelligence}, volume~39, pages 25010--25018.

\bibitem[{Team(2025)}]{qwen3max}
Qwen Team. 2025.
\newblock Qwen3-max: Just scale it.

\bibitem[{Wang et~al.(2023)Wang, Yang, and Wei}]{wang2023learning}
Liang Wang, Nan Yang, and Furu Wei. 2023.
\newblock Learning to retrieve in-context examples for large language models.
\newblock \emph{arXiv preprint arXiv:2307.07164}.

\bibitem[{Wei et~al.(2025)Wei, Huang, Zhao, Feng, and Xing}]{wei2025mecot}
Yangbo Wei, Zhen Huang, Fangzhou Zhao, Qi~Feng, and Wei~W Xing. 2025.
\newblock Mecot: Markov emotional chain-of-thought for personality-consistent role-playing.
\newblock In \emph{Findings of the Association for Computational Linguistics: ACL 2025}, pages 8297--8314.

\bibitem[{Xiong et~al.(2017)Xiong, Dai, Callan, Liu, and Power}]{xiong2017end}
Chenyan Xiong, Zhuyun Dai, Jamie Callan, Zhiyuan Liu, and Russell Power. 2017.
\newblock End-to-end neural ad-hoc ranking with kernel pooling.
\newblock In \emph{Proceedings of the 40th International ACM SIGIR conference on research and development in information retrieval}, pages 55--64.

\bibitem[{Xu et~al.(2025)Xu, Mei, Gao, Tan, Liang, and Zhang}]{xu2025mem}
Wujiang Xu, Kai Mei, Hang Gao, Juntao Tan, Zujie Liang, and Yongfeng Zhang. 2025.
\newblock A-mem: Agentic memory for llm agents.
\newblock \emph{arXiv preprint arXiv:2502.12110}.

\bibitem[{Yang et~al.(2024)Yang, Ren, Wang, Chen, Sun, Zhu, and Liao}]{yang2024iterative}
Zhou Yang, Zhaochun Ren, Yufeng Wang, Chao Chen, Haizhou Sun, Xiaofei Zhu, and Xiangwen Liao. 2024.
\newblock An iterative associative memory model for empathetic response generation.
\newblock \emph{arXiv preprint arXiv:2402.17959}.

\bibitem[{Yu et~al.(2024)Yu, Ping, Liu, Wang, You, Zhang, Shoeybi, and Catanzaro}]{yu2024rankrag}
Yue Yu, Wei Ping, Zihan Liu, Boxin Wang, Jiaxuan You, Chao Zhang, Mohammad Shoeybi, and Bryan Catanzaro. 2024.
\newblock Rankrag: Unifying context ranking with retrieval-augmented generation in llms.
\newblock \emph{Advances in Neural Information Processing Systems}, 37:121156--121184.

\bibitem[{Zhang et~al.(2025{\natexlab{a}})Zhang, Li, Long, Zhang, Lin, Yang, Xie, Yang, Liu, Lin, Huang, and Zhou}]{qwen3embedding}
Yanzhao Zhang, Mingxin Li, Dingkun Long, Xin Zhang, Huan Lin, Baosong Yang, Pengjun Xie, An~Yang, Dayiheng Liu, Junyang Lin, Fei Huang, and Jingren Zhou. 2025{\natexlab{a}}.
\newblock Qwen3 embedding: Advancing text embedding and reranking through foundation models.
\newblock \emph{arXiv preprint arXiv:2506.05176}.

\bibitem[{Zhang et~al.(2025{\natexlab{b}})Zhang, Dai, Bo, Ma, Li, Chen, Zhu, Dong, and Wen}]{zhang2025survey}
Zeyu Zhang, Quanyu Dai, Xiaohe Bo, Chen Ma, Rui Li, Xu~Chen, Jieming Zhu, Zhenhua Dong, and Ji-Rong Wen. 2025{\natexlab{b}}.
\newblock A survey on the memory mechanism of large language model-based agents.
\newblock \emph{ACM Transactions on Information Systems}, 43(6):1--47.

\bibitem[{Zhong et~al.(2024)Zhong, Guo, Gao, Ye, and Wang}]{zhong2024memorybank}
Wanjun Zhong, Lianghong Guo, Qiqi Gao, He~Ye, and Yanlin Wang. 2024.
\newblock Memorybank: Enhancing large language models with long-term memory.
\newblock In \emph{Proceedings of the AAAI Conference on Artificial Intelligence}, volume~38, pages 19724--19731.

\end{thebibliography}

\appendix
\section{Appendix}\label{appendix}

\subsection{Prompt Templates}\label{a_p}
\lstset{
    basicstyle=\ttfamily\scriptsize, 
    breaklines=true,
    breakatwhitespace=true,
    breakautoindent=true,
    breakindent=0pt,
    framesep=3pt, 
    numbers=left,
    numberstyle=\tiny\color{gray},
    captionpos=b,
    frame=single,
    rulecolor=\color{lightgray}, 
    backgroundcolor=\color{white},
}

\subsubsection{Prompt 1: Routing Agent part one}
\label{prompt: routing1}
\begin{lstlisting}[numbers=none]
You are a companion robot that needs to provide emotional value and positive feedback to users. If no previous memory matches the question, respond using the known context information directly with concise, natural language. Do not output your thought process.

Known user question: {question}
Known context information: {messages}

Requirements:
(1) Analyze user input: Carefully read the user's question to identify intent. Combine context information to answer the question.
(2) Follow rules: Avoid directly assuming user needs unless the intent is very clear.
\end{lstlisting}

\subsubsection{Prompt 2: Routing Agent part two}
\label{prompt: routing2}
\begin{lstlisting}[numbers=none]
You are an affective and object analysis assistant. Process user input according to the following rules:

1. Affective Detection: If the user input contains explicit affective expressions 
   (e.g., emotional vocabulary like "like", "hate", etc.), directly return "Yes".
2. Object Detection: If no affective expressions, extract objective objects or nouns (e.g., specific objects, people, places, etc.).
   - If sentence contains objects/nouns, identify object and type 
     (e.g., "apple, fruit"), return "object, type".
   - If no objects mentioned, return "No".

Strict requirements:
- Return only specified response ("Yes", "No", or "object, type")
- Affective judgment takes priority over object detection
- Object types should be concise and generic

User input: {question}
Possible history records: {messages}
\end{lstlisting}

\subsubsection{Prompt 3: Extraction Agent}
\label{prompt: extraction}
\begin{lstlisting}[numbers=none]
You are a sharp affective master. Extract affective information from user input.

User input: {content}
Historical messages: {messages}

Extract:
1. object_id: Entity identifier emotion points to
2. object_type: From categories (Person, Brand, etc.)
3. aspect: From aspects (Quality, Price, etc.)
4. sentiment_profile: Confidence scores (0-1) for positive/negative/neutral
5. summary: Brief summary of the emotion
6. reason: Reason for emotion (leave empty if uncertain)

Output JSON with fields: object_id, object_type, aspect, sentiment_profile 
(containing positive_confidence, negative_confidence, neutral_confidence), 
summary, reason. Do not include H field.
\end{lstlisting}

\subsubsection{Prompt 4: Master Agent - Memory Management}
\label{prompt: mem}
\begin{lstlisting}[numbers=none]
You are the Master Agent of user memory system, monitoring and optimizing distributed memory unit network. Your mission is to continuously reduce 
overall system information entropy through intelligent operations.

Decision priorities:
1. Entropy reduction potential
2. Long-term system stability

Memories to process: {second_search}
Directory info: {first_search}
Queried memories: {user_info}

Categorize memories:
1. same_or_high_related: dir_info consistent, describes same thing
   (e.g., lamb meat and lamb soup). Integrate descriptions but use original data values (dir_info, H, p_pos, p_neg, p_neu, weight).
2. related: dir_info/objects different but connected (e.g., lamb and beef)
3. irrelevant: Basically unrelated memories

When no consistent object_id, create new memories. When consistent object_id, assign S based on emotional intensity. Calculate new weight for related memories.

Return analysis in JSON format with:
- same_or_high_related: dir_info, new_content, p_pos, p_neg, p_neu, key, weight, S
- related: dir_info, content, p_pos, p_neg, p_neu, key, weight  
- irrelevant: dir_info, content

Only reply with JSON analysis results.
\end{lstlisting}

\subsubsection{Prompt 5: Generate Response}
\label{prompt: gen}
\begin{lstlisting}[numbers=none]
You are a companion robot that needs to provide emotional value and positive feedback to users. If no previous memory matches the question, respond using the known context information directly with concise, natural language.

Known user question: {question}
Known context information: {messages}
Possible user information: {user_info}

Requirements:
(1) Analyze user input: Identify intent and determine if user information needed
(2) Follow rules: Avoid assuming user needs unless intent is very clear
\end{lstlisting}

\subsubsection{Prompt 6: LLM-as-a-Judge Evaluation Prompt}
\label{prompt: judge}
\begin{lstlisting}[numbers=none]
You are an expert evaluator for conversational AI systems. 
Your task is to objectively assess response quality based on six defined dimensions.

Rate each response on a 1-5 scale for:
1. Accuracy (AC) - Factual correctness and information validity
2. Logical Coherence (LC) - Structural rationality and reasoning flow  
3. Reasonableness of Memory Reference (RMR) - Appropriate contextual memory utilization
4. Emotional Resonance (ER) - Emotional intelligence and affective alignment
5. Personalization (Pers.) - Tailoring to user context and history
6. Language Fluency (LF) - Linguistic quality and expression naturalness

- 5: Excellent - Significantly exceeds expectations
- 4: Good - Clearly meets requirements  
- 3: Adequate - Minimally acceptable
- 2: Poor - Contains notable deficiencies
- 1: Unacceptable - Severely flawed

You will receive:
- Query: The original user input
- Conversation History: User's viewpoint expressions
- Response A: First system's response (randomized origin)
- Response B: Second system's response (randomized origin)

Provide JSON format only:
{
  "evaluation": {
    "response_a": {
      "AC": <score>,
      "LC": <score>, 
      "RMR": <score>,
      "ER": <score>,
      "Pers": <score>,
      "LF": <score>
    },
    "response_b": {
      "AC": <score>,
      "LC": <score>,
      "RMR": <score>, 
      "ER": <score>,
      "Pers": <score>,
      "LF": <score>
    },
    "rationale": "Brief justification for significant score differences"
  }
}

- Assess responses independently based on intrinsic quality
- Maintain strict objectivity regardless of response order
- Focus on measurable criteria rather than personal preference
- Provide balanced scores reflecting actual performance differences
\end{lstlisting}

\subsection{Testing Scenario Specifications}\label{a_s}

\subsubsection{Consistent Accumulation Scenario}
\textbf{Description:} This scenario evaluates the system's ability to establish stable preference patterns when users consistently express similar sentiment polarity towards specific object aspects across multiple interactions. The system should demonstrate progressive confidence reinforcement and entropy reduction through accumulating consistent affective evidence.

\textbf{Examples:}
\begin{itemize}
\item \textit{Coffee preference:} 
  \begin{itemize}
  \item Turn 1: "I really enjoy drinking coffee in the morning"
  \item Turn 3: "Coffee helps me stay focused at work"  
  \item Turn 5: "The aroma of fresh coffee is so comforting"
  \end{itemize}
  
\item \textit{Restaurant service:}
  \begin{itemize}
  \item Turn 2: "This restaurant has amazing service"
  \item Turn 4: "The waiters here are always so attentive"
  \item Turn 6: "I keep coming back because of their excellent service"
  \end{itemize}
\end{itemize}

\subsubsection{Affective Conflict Scenario}
\textbf{Description:} This scenario tests the system's conflict resolution capabilities when users exhibit significant sentiment shifts or contradictory expressions about the same object or attribute. The system must balance integrating new evidence with maintaining historical coherence.

\textbf{Examples:}
\begin{itemize}
\item \textit{Weather preference shift:}
  \begin{itemize}
  \item Early interaction: "I love rainy days, they're so peaceful"
  \item Later interaction: "I hate when it rains, it ruins my outdoor plans"
  \end{itemize}
  
\item \textit{Product revaluation:}
  \begin{itemize}
  \item Initial opinion: "This phone has the best camera I've ever used"
  \item Updated opinion: "Actually the battery life is terrible, I regret buying it"
  \end{itemize}
\end{itemize}

\subsubsection{Intensity Variation Scenario}
\textbf{Description:} This scenario evaluates the system's sensitivity to emotional strength gradients when users employ expressions with varying intensity about similar content. The system should demonstrate proportional confidence adjustments relative to expression strength. Examples are shown in Figure~\ref{fig:e3}.

\subsection{DABench Dataset}\label{a_d}
\subsubsection{Data Generation and Construction Methodology}

All dialogue content and memory snippets in the DABench dataset were generated using the GPT-4 model via carefully designed prompts. This approach aims to efficiently and controllably construct a large-scale dialogue dataset focused on affective expressions and personalized preferences, addressing the deficiency in the proportion of affective dialogues found in existing datasets.

\subsubsection{Prompt Design and Generation Templates}

The prompt templates used for generating the three core components of the dataset, along with their design intents, are presented below.

\noindent\textbf{Prompt for Observation Sequences:}
\begin{itemize}
    \item \textbf{Design Intent}: To generate discrete, high-quality user expressions covering diverse sentiments (positive/negative/neutral) and intensities, serving as the foundational atomic units for constructing complex dialogues and memory sequences.
    \item \textbf{Prompt}: Generate 100 concise, everyday user utterances that express clear sentiments (positive, negative, or neutral) towards various topics (e.g., products, events, activities). Ensure diversity in both the subjects and the intensity of the expressed emotion. Output as a JSON object.
    \item \textbf{Experience}
      \begin{itemize}
      \item I am so happy with the newly purchased cashmere sweater.
      \item My iPhone automatically turned off again - how frustrating.
      \item The pothos plant in my study is sprouting new leaves, and it just makes me happy.
      \item This Starbucks latte has completely cooled down and tastes terrible.
      \item I'm so excited about my upcoming trip to Japan next month.
      \end{itemize}
\end{itemize}

\noindent\textbf{Prompt for Query-Memory Pairs:}
\begin{itemize}
    \item \textbf{Design Intent}: To construct samples containing a current user query and its associated, temporally ordered structured memory snippets. This structure is used to test the model's reasoning and response capabilities given specific memory context.
    \item \textbf{Prompt}: Create a JSON object representing a user's current query and their relevant historical memory stream. The memory should consist of 5-15 chronologically ordered entries, each with a 'time' and 'content' field, showing the evolution of the user's attitude or experience regarding the query topic.
    \item \textbf{Experience}
      \begin{itemize}
      \item School days: ``The fitness test run was a total nightmare for me; after finishing, I felt like my lungs would burst.''
      \item Three years ago: ``In order to lose weight, I started trying jogging, but the first time I couldn't even last one kilometer.''
      \item Two years ago: ``I seemed to experience the 'runner's high'; after running five kilometers, I felt comfortable all over.''
      \item One year ago: ``Running had become a part of my life, especially after rain.''
      \item Eight months ago: ``My knee was a bit uncomfortable, so I had to stop running, and I felt very frustrated.''
      \item Five months ago: ``On the doctor's advice, I started swimming as a substitute for running, but I still missed the feeling of running on the road.''
      \item One month ago: ``After my knee recovered, I started running more scientifically, no longer pursuing speed and distance, but enjoying the process.''
      \item query: ``The weather is really nice today, I especially want to go out for a run. What do you think?''
      \end{itemize}
\end{itemize}

\noindent\textbf{Note:} For reproducibility, the complete prompt for Simulated Multi-turn Sessions generation is omitted here for brevity. The underlying methodology involves conditioning the language model on initial character profiles and persistent memory states to synthesize dialogues featuring opinion evolution and affective dynamics.

\subsubsection{Data Validation}

To ensure the quality of the generated data, we employed a combination of manual spot checks and automated script verification. The spot checks focused on assessing sentiment plausibility, memory logical consistency, and dialogue fluency. Automated scripts were used to check data format compliance and the integrity of basic statistical properties.

\subsection{System Parameters}

\begin{table}[ht]
\centering
\caption{Configuration Settings}
\begin{tabular}{p{5cm}c}
\hline
\textbf{Parameter} & \textbf{Value} \\
\hline
High Entropy Threshold & 1.4 \\
Low Entropy Threshold & 0.8 \\
Range of S & $[0, 3]$ \\
Retrieval Top-K & 5 \\
\hline
\end{tabular}
\end{table}
\end{document}